\documentclass{article}

\PassOptionsToPackage{numbers, compress}{natbib}

\usepackage[main, final]{neurips_2026}

\usepackage[utf8]{inputenc} 
\usepackage[T1]{fontenc}    
\usepackage{hyperref}       
\usepackage{url}            
\usepackage{booktabs}       
\usepackage{amsfonts}       
\usepackage{nicefrac}       
\usepackage{microtype}      
\usepackage{xcolor}         
\usepackage{graphicx}
\usepackage{amssymb}
\usepackage{algorithm}
\usepackage{algpseudocode}
\usepackage{amsmath}        
\usepackage[capitalize]{cleveref}
\usepackage{wrapfig}
\usepackage{float}
\usepackage{caption}
\usepackage{multirow}
\usepackage{booktabs}
\usepackage{enumitem}
\usepackage{graphicx}      
\usepackage{amsmath}       
\usepackage{multirow}      
\usepackage{booktabs}      
\usepackage[table]{xcolor}
\usepackage{array}
\title{LiWi: Layering in the Wild}

\author{
    \textbf{Yu He}\textsuperscript{1}\thanks{Equal contribution} ,
    \textbf{Fang Li}\textsuperscript{1}\footnotemark[1] ,
    \textbf{Haoyang Tong}\textsuperscript{1,2} ,
    \textbf{Lichen Ma}\textsuperscript{1} ,
    \textbf{Xinyuan Shan}\textsuperscript{1} ,
    \textbf{Jingling Fu}\textsuperscript{1}\\
    \textbf{Dong Chen}\textsuperscript{1} ,
    \textbf{Luohang Liu}\textsuperscript{1} ,
    \textbf{Junshi Huang}\textsuperscript{1}\thanks{Corresponding Author.} ,
    \textbf{Yan Li}\textsuperscript{1} \\
    \textsuperscript{1}JD.com \quad
    \textsuperscript{2}MAIS \& NLPR, CASIA \\
    {\tt\small \{heyu2579, junshi.huang\}@gmail.com}
}

\begin{document}

\maketitle

\begin{abstract}
    Recent advances in generative models have empowered impressive layered image generation, yet their success is largely confined to graphic design domains.
    The layering of in-the-wild images remains an underexplored problem, limiting fine-grained editing and applications of images in real-world scenarios.
    Specifically, challenges remain in scalable layered data and the modeling of object interaction in natural images, such as illumination effects and structural boundary. 
    To address these bottlenecks, we propose a novel framework for high-fidelity natural image decomposition. 
    First, we introduce an \emph{Agent-driven Data Decomposition} (ADD) pipeline that orchestrates agents and tools to synthesize layered data without manual intervention. 
    Utilizing this pipeline, we construct a large-scale dataset, named LiWi-100k, with over 100,000 high-quality layered in-the-wild images.
    Second, we present a novel framework that jointly improves photometric fidelity and alpha boundary accuracy.
    Specifically, shadow-guided learning explicitly models the illumination effects, and degradation-restoration objective provides boundary-correction supervision by recovering clean foreground image from degraded one.
    Extensive experiments demonstrate that our framework achieves state-of-the-art (SoTA) performance in natural image decomposition, outperforming existing models in RGB L1 and Alpha IoU metrics.
    We will soon release our code and dataset.
    \vspace{2mm} 
    \noindent \textbf{Project Page:} \url{https://rassetmusty.github.io/LiWi/}
\end{abstract}

\section{Introduction}

Layer decomposition aims to convert a flattened image into a set of visual elements, such as foreground objects with their alpha masks and a clean background. 
It unlocks essential structural priors required for controllable video generation (\textit{e.g.}, independent foreground-background motion), 3D asset synthesis (\textit{e.g.}, cues for occluded entities), and the development of interactive world models~\cite{kasten2021layered,bartal2022text2live,lee2023shapeaware}. 
Compared with conventional segmentation or matting, image layering requires not only identifying visible object regions but also recovering complete layer appearances and the scene content behind them~\cite{suvorov2022lama}. 
This makes in-the-wild image layering a useful intermediate representation between pixel-level image generation and structured visual understanding.

Despite recent progress in layered image generation and decomposition, most existing methods~\cite{suzuki2025layerd,yin2025qwen,liu2025omnipsd} are mostly compatible to the decomposition of graphic design, PSD (Photoshop Document) assets, or synthetically composed images.
These domains usually contain clean boundaries, explicit layer ordering, and simple alpha blending. 
Real photographs are more challenging. 
Foreground objects do not merely occlude the background; they also change the scene through cast shadows, contact darkening, reflections, soft boundaries, and local illumination variations~\cite{garces2022survey}. 
The target of natural image layering is not only to separate visible elements but also to decide where the physical traces caused by those objects should go.
As a result, a real-world image cannot be fully explained by simply stacking RGBA layers. 

A central obstacle of natural image decomposition is the lack of training data for in-the-wild image layering task. 
Unlike graphic designs, real-world images do not provide authored layers. 
Manually annotating is expensive and difficult to scale. 
To address this data bottleneck, we propose the ADD pipeline that constructs layered supervision from in-the-wild images without manual annotation. 
ADD enables agents and specialized tools to generate clean backgrounds, complete foreground RGBA layers, and select consistent layer combinations.

However, high-quality layered data alone is not sufficient for natural photographs. 
Real-world scenes are governed by complex illumination. 
Effects such as shadows and lighting variations are contextual footprints induced by foreground objects and background scenes. 
We therefore introduce a shadow layer to explicitly represent this photometric residual between the target image and the recomposed image. 
Instead of forcing such residuals to be ambiguously absorbed by the foreground or background, the shadow layer provides supervision for global illumination interactions. 
This encourages the model to disentangle visual traces induced by foreground entities.

Beyond color fidelity, in-the-wild image layering also requires accurate layer boundaries.
We observe that many failure cases arise from local boundary degradation, including mask erosion, slight dilation,  and inaccurate color blending near object contours.
To address these boundary-level errors, we introduce a degradation-restoration objective as an auxiliary foreground refinement task. 
During training, foreground layers are deliberately corrupted, and the model is trained to recover the corresponding clean layers. 
This restoration-oriented supervision encourages the model to capture the mechanisms behind alpha boundary formation, local color correction, and texture preservation. 

Our main contributions are summarized as follows:
\begin{itemize}[itemsep=2pt, topsep=0pt, parsep=0pt]
    \item We propose the ADD pipeline and construct LiWi-100k, a large-scale and high-quality dataset for in-the-wild image layering, eliminating the need for expensive manual annotation.
    \item We propose a layer decomposition framework that combines the shadow layer with auxiliary layer refinement. The shadow residual captures photometric variations, while the degradation-restoration objective improves boundary accuracy.
    \item Extensive experiments demonstrate that our framework achieves SoTA performance both on LiWi-100k and Crello \cite{yamaguchi2021canvasvae}, outperforming existing models in RGB L1 and Alpha IoU.
\end{itemize}

\section{Related Work}

\subsection{Image Layer Decomposition} 
Layered image decomposition provides an interpretable representation for image editing, compositional generation, and inverse graphics~\cite{huang2018deepprimitive,yang2025generative}. Recent work has advanced from synthetic compositions to editable full-RGBA representations \cite{zhang2023text2layer,huang2024layerdiff,pu2025art}, including matting-based data construction in Text2Layer~\cite{zhang2023text2layer}, modular open-domain decomposition in MULAN~\cite{tudosiu2024mulan}, iterative top-layer extraction for graphic designs in LayerD~\cite{suzuki2025layerd}, and end-to-end diffusion-based RGB-to-RGBA decomposition in Qwen-Image-Layered~\cite{yin2025qwen,wu2025qwen}. However, natural-image decomposition remains difficult: object-centric pipelines accumulate errors across intermediate modules~\cite{tudosiu2024mulan}, design-oriented methods assume clean boundaries and organized layers rarely found in photographs~\cite{suzuki2025layerd,chen2025rethinking}, and recent end-to-end approaches are trained on PSD-like authoring data, making them better suited to design-style semantic layers than to natural-scene photometry~\cite{yin2025qwen,liu2025omnipsd}. Real photographs involve entangled shadows, reflections, translucency, soft transitions, and occlusions, which complicate both layer separation and cross-layer interaction modeling~\cite{yang2026controllable,chen2026referring,yang2025generative}. We address this gap by decomposing natural images with a training strategy that better preserves photometric effects and compositional interactions.

\subsection{RGBA Dataset Construction} 
Training data for layered image modeling typically follows two routes: synthetic composition, which composites foregrounds, masks, or transparent layers under predefined blending rules to provide scalable multilayer supervision with explicit control over layer order and alpha blending~\cite{zhang2024transparent,huang2025dreamlayer,huang2026psdiffusion}; and extraction-based pipelines, which derive foregrounds from segmentation or matting, reconstruct backgrounds via inpainting, and infer layer order from geometric or learned cues~\cite{kang2025layeringdiff,yang2025generative}. However, both remain insufficient for natural-image layer decomposition. Synthetic data often exhibits overly clean interactions and a realism gap~\cite{dalva2024layerfusion,chen2025inpainting}, whereas extraction-based pipelines are vulnerable to upstream errors and accumulate structural and photometric artifacts across stages~\cite{kang2025layeringdiff}. Agent-style automation can improve scalability, but tightly coupled multi-stage workflows remain brittle when multiple dependent decisions must be jointly correct~\cite{chen2026referring,sun2025datasetagent}. To address these limitations, we propose a decoupled data construction pipeline that separately builds backgrounds, foregrounds, and final layered composites, reducing inter-stage interference while preserving layered consistency and photometric realism. A consensus-based verification mechanism further filters unreliable samples, enabling a more scalable and reliable dataset for natural-image layer decomposition.

\section{Synthesizing Layered Images in the Wild}

\begin{figure*}[t]
    \centering
    \includegraphics[width=\textwidth]{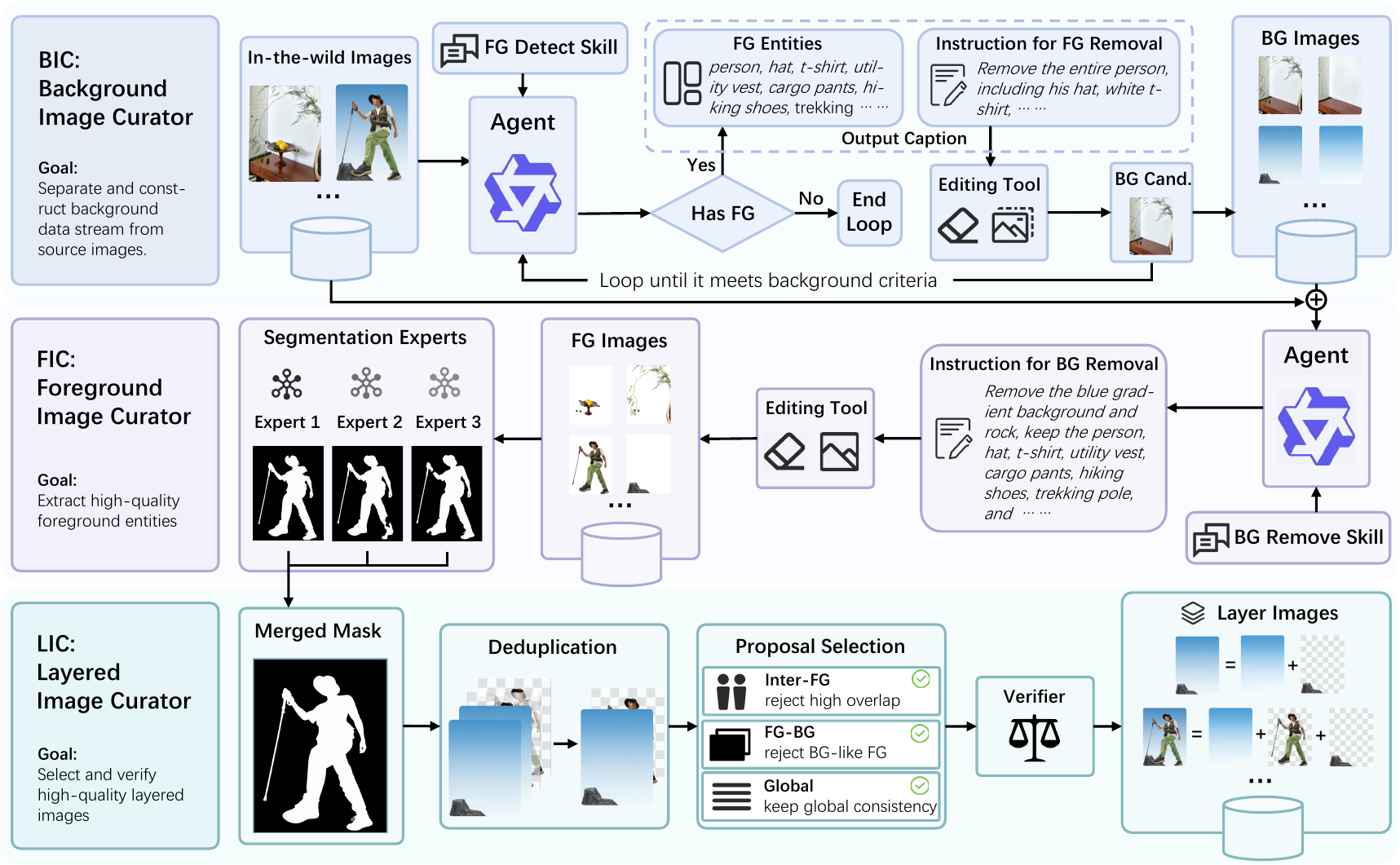}
    \caption{Overview of our ADD pipeline. The system leverages agent and specialized tools to automatically decompose in-the-wild images. 
    Foreground and background layers are routed into separate repositories and subsequently selected by the LIC module, where a rigorous verifier ensures the quality of the final layered compositions.}
    \label{fig:data_pipe}
    \vspace{-10pt}
\end{figure*}



Learning in-the-wild image layering requires supervision that is rarely available in real photographs. 
Unlike graphic designs or PSD files, where layers are explicitly authored, an in-the-wild image only provides a flattened RGB observation in which foreground appearance, occlusion, cast shadows, reflections, and illumination changes are entangled. 
A simple segmentation mask can recover the visible foreground region, but it does not reveal the clean background behind the object, nor does it explain the photometric footprint left by the foreground on the scene. 
To address these problems in layering task, we introduce the ADD pipeline, a multi-agent system that automatically synthesizes high-quality layered samples from in-the-wild images.

\subsection{Problem Formulation and Overview}

Given a collection of in-the-wild images $\mathcal{I}$, our goal is to construct a layered dataset
$\mathcal{D}=\{(I_{src}, B, \{F_k, \alpha_k\}_{k=1}^{K})\}$,
where $I_{src}$ is the input image to be decomposed into background image $B$ and foreground images $\{F_k, \alpha_k\}_{k=1}^{K}$.
$F_k$ and $\alpha_k$ denote the RGB appearance and alpha mask of the $k$-th foreground image. 
Note that $I_{src}$ can be original image from $\mathcal{I}$ or intermediate background in data curation.
The key requirement of layered images is that all components should be both individually valid and jointly consistent.
Specifically, foreground entities should be complete and semantically meaningful, background should be free of foreground artifacts, and their composition $I_{src}$ should preserve plausible spatial and photometric interactions.

As shown in \cref{fig:data_pipe}, the proposed ADD is implemented as an agentic system and contains three collaborative curators: the \emph{Background Image Curator} (BIC), the \emph{Foreground Image Curator} (FIC), and the \emph{Layered Image Curator} (LIC). 
BIC builds a repository of clean backgrounds, FIC extracts high-quality foreground entities with transparent masks, and LIC selects compatible foreground-background combinations to produce final layered samples. 
This agent-driven mechanism enables scalable data construction while avoiding the requirement for manual intervention.


\subsection{Background and Foreground Curation}

Given an image $I \in \mathcal{I}$, the BIC constructs a pool of background candidates $\mathcal{B}$ in a loop starting from $I_0 = I$. 
In the $i$-th ($i \ge 0$) step, the agent first detects whether there is foreground entity in the input $I_{i}$ by foreground detection skill.
If no foreground is detected, the loop ends.
Otherwise, the agent generates an editing instruction that describes the complete foreground region to be removed, including the main object, accessories, and visually attached parts. 
The agent then calls an editing tool to produce a background candidate $B_{i+1}$, which is set to $I_{i+1}$ for the next step, based on the foreground removal instruction.
Note that the foreground descriptions are reusable in FIC.

Based on raw image $I$ and background candidates $\mathcal{B}$, the FIC builds a foreground repository $\mathcal{F}$ containing complete foreground entities.
Rethinking the $i$-th step of BIC, the dominated foreground entities can be detected in the input image $I_i$ where $i \in \{0, 1, ..., |\mathcal{B}|-1\}$.
With the detected foreground entities, the agent generates a background removal instruction by specifying the retained foreground entities and all visually attached components.
The editing tool then erases the surrounding background and produces a foreground image with white background, denoted as $\tilde{F}_{i+1}$, for simple segmentation.
Since a single segmentation may produce incomplete mask around the regions of thin structures, accessories, or boundaries, we use $N$ segmentation experts to obtain candidate masks $\{M^{(1)}_{i+1}, M^{(2)}_{i+1}, \ldots, M^{(N)}_{i+1}\}$ from $\tilde{F}_{i+1}$.
We merge these candidate masks to generate $\alpha_{i+1} = \mathrm{avg}(M^{(1)}_{i+1}, M^{(2)}_{i+1}, \ldots, M^{(N)}_{i+1})$.
The final RGBA foreground image $F_{i+1}$ is constructed by merging the RGB content of $\tilde{F}_{i+1}$ with the alpha map $\alpha_{i+1}$.
In this way, the foreground extraction is simplified by first removing the complex background context and then using multi-expert mask fusion to obtain a complete alpha map.

\subsection{Layered Composition and Verification}

Given the background pool $\mathcal{B}$ produced by BIC and the RGBA foreground pool $\mathcal{F}$ produced by FIC, LIC de-duplicates the near-identical images within $\mathcal{B}$ and $\mathcal{F}$ respectively. 
This avoids excessive visual redundancy in the selection process of foreground-background compositions.
In practice, we use DINOv2 embedding to represent images.

\begin{figure}[t]
\centering

\begin{minipage}[t]{0.48\textwidth}
\vspace{0pt}

\hrule height 0.8pt
\vspace{3pt}

\captionsetup{
    type=algorithm,
    justification=raggedright,
    singlelinecheck=false,
    font=small,
    labelfont=bf,
    skip=0pt
}
\captionof{algorithm}{Proposal Selector for LIC}
\label{alg:selector}

\vspace{3pt}
\hrule height 0.4pt
\vspace{3pt}

\scriptsize
\begin{algorithmic}[1]
\Require input image $I$; backgrounds $\mathcal{B}=\{B_k\}_{k=1}^{K}$;
foregrounds $\mathcal{F}=\{(\tilde{F}_k,\alpha_k)\}_{k=1}^{K}$;
DINOv2 $\phi(\cdot)$; thresholds $\tau_{local},\tau_{global}$.
\Ensure proposals of layered images $\mathcal{P}$.

\State $M(\alpha,X)\triangleq X\odot\alpha+\mathbf{1}(1-\alpha)$ \Comment{Function Definition}
\State $f_{ij}^{F}\gets\phi(M(\alpha_i,\tilde{F}_j)), \forall i,j \in [1, 2, ..., K]$
\State $f_{ij}^{B}\gets\phi(M(\alpha_i,B_j)), \forall i,j \in [1,2,...,K]$
\State $\mathcal{P}\gets\varnothing$, $\mathcal{F}_{valid} \gets \varnothing$

\For{$\mathcal{F}_{sub} \subseteq \mathcal{F}$} \Comment{Inter-FG Overlap}
    \State \textbf{if} $\exists F_i, F_j \in \mathcal{F}_{sub}, i < j$ and $\langle f_{ii}^F,f_{ij}^{F}\rangle>\tau_{local}$
        \State \quad \textbf{continue}
    \State \textbf{else}
        \State \quad $\mathcal{F}_{valid} \gets \mathcal{F}_{sub} \cup \mathcal{F}_{valid}$
\EndFor

\For{$I_{src} \in \{I\}\cup\mathcal{B},\ B_j \in \mathcal{B}\setminus\{I_{src}\},\ \mathcal{F}_{sub} \subseteq \mathcal{F}_{valid}$}
    \State \textbf{if} $\exists F_i\in \mathcal{F}_{sub}, \langle f_{ii}^F,f_{ij}^{B}\rangle>\tau_{local}$ \Comment{FG-BG Overlap}
        \State \quad \textbf{continue}

    \State $I_c\gets\textsc{Composite}(B_j, \mathcal{F}_{sub})$

    \State \textbf{if} $\langle\phi(I_c), \phi(I_{src}) \rangle \ge \tau_{global}$ \Comment{Global Consistency }
    
        \State \quad $\mathcal{P}\gets\mathcal{P}\cup \{(I_{src},B_j,\mathcal{F}_{sub})\}$
\EndFor

\State \Return $\mathcal{P}$
\end{algorithmic}

\vspace{3pt}
\hrule height 0.8pt

\end{minipage}
\hfill
\begin{minipage}[t]{0.48\textwidth}
\vspace{-3pt}
\centering
\includegraphics[width=\linewidth]{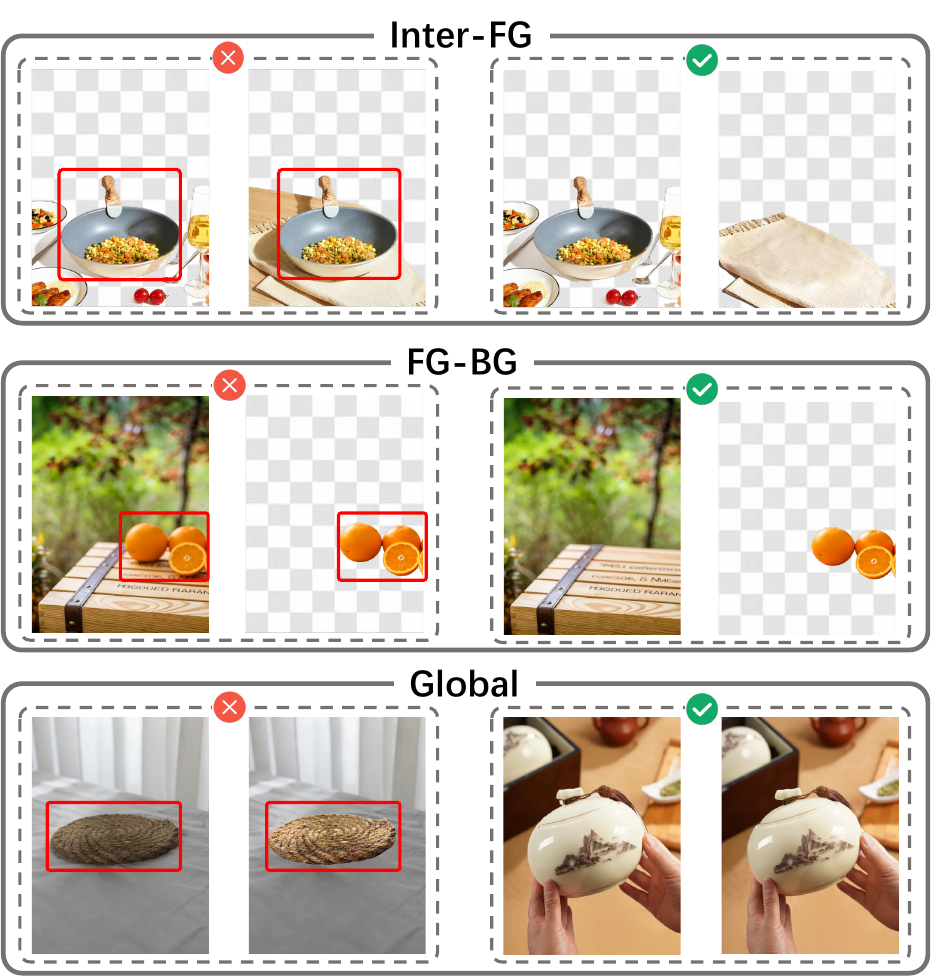}

\vspace{3pt}
\captionsetup{
    type=figure,
    justification=centering,
    singlelinecheck=false,
    font=small,
    skip=2pt
}
\captionof{figure}{Illustration of pass and fail examples from Inter-FG, FG-BG and Global Consistency constraints.}
\label{fig:proposal}
\end{minipage}

\vspace{-6pt}
\end{figure}

After de-duplication, LIC selects the compatible foreground-background compositions by verifying the full combinations of candidates.
As summarized in Alg.~\ref{alg:selector} and illustration of passed and failure examples in \cref{fig:proposal}, the Proposal Selector evaluates every combination from three perspectives. 
First, the \emph{Inter-FG} constraint removes foreground combinations with strong overlap or semantic redundancy. 
For each foreground pair $(F_i,F_j)$ with $i < j$ (\textit{i.e.}, $F_i$ is in front of $F_j$ and can occlude it), we use the mask of $F_i$ (\textit{i.e.}, $\alpha_i$) to cut out the corresponding region of $F_j$.
If the similarity is abnormally high, the two layers are likely to describe the same object or heavily occluded regions, and the candidate is rejected. 
Second, the \emph{FG-BG} constraint checks whether a background contains the foreground contents. 
If the masked background region is highly similar to the foreground itself, the background is likely to retain redundant elements and should be discarded. 
For proposals that pass the entity-level checks, the \emph{Global Consistency} constraint evaluates the rendered composition $I_c$. 
The holistic consistency of $I_c$ is then measured against source images $I_{src}$. 
Only compositions that maintain a high global similarity score are retained, ensuring that the selected layers form a plausible natural image rather than an arbitrary collage.

The remaining composition candidates are further examined by a model verifier. 
The verifier checks both layer-wise quality and composition-level validity, including foreground completeness, background cleanliness, absence of obvious artifacts, and the semantic plausibility of the rendered image. 
Candidates that fail these checks are rejected, while accepted candidates are stored as final layered samples.
Through this selector-verifier design, LIC automatically transforms independent background and foreground streams into consistent layered images, providing scalable, human-free supervision for in-the-wild images.

\subsection{LiWi-100k Dataset}

\begin{wrapfigure}{r}{0.54\textwidth}
    \vspace{-12pt}
    \centering
    \includegraphics[width=0.53\textwidth]{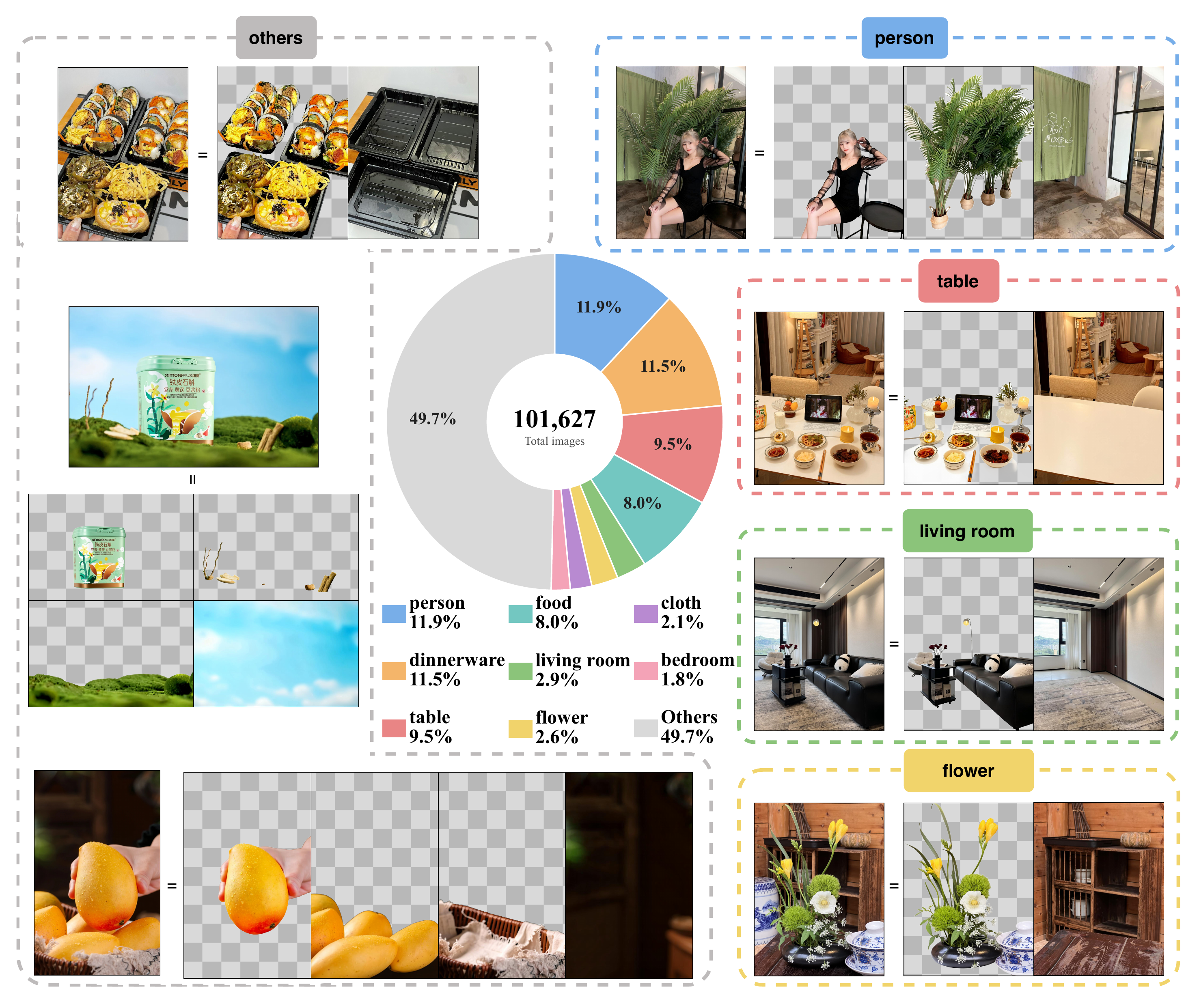}
    \caption{Data distribution and samples of LiWi-100k.}
    \label{fig:dataset}
    \vspace{-12pt}
\end{wrapfigure}

We introduce LiWi-100k, a layered dataset dedicated exclusively to real-world scenes. 
It contains 101,627 high-quality layered images built entirely from unstructured real-world images without manual annotation.
The curation process is fully automated through our proposed ADD pipeline, which orchestrates a suite of open-source models. 
The Agent and Verifier are instantiated with Qwen3-VL-32B \cite{bai2025qwen3}, generating precise removal instructions and verifying the proposals. 
The Editing Tool is powered by FLUX.2-klein-9B \cite{bfl_flux2_klein_2026}, ensuring high-fidelity removal for background and foreground. 
For segmentation, we employ an ensemble of experts comprising RMBG-1.4 \cite{bria_rmbg14_2024}, RMBG-2.0 \cite{BiRefNet, bria_rmbg2_2024}, and SAM3 \cite{carion2025sam}. 


As illustrated in \cref{fig:dataset}, LiWi-100k encompasses a broad categories of images in real-world scenarios, ensuring rich compositional diversity. 
Regarding structural complexity, the dataset contains a maximum of 5 layers. 
The vast majority of the samples (89\%) consist of 2 layers, while the remaining 11\% contain 3 to 5 layers. 
Unlike graphic designs, where numerous individual visual elements are artificially stacked, real-world photographs typically center around one or two primary subjects interacting with a holistic environment. 
The complexity in natural scene images lies not in the sheer quantity of layers, but in the intricate physical entanglement between foreground and scene, such as cast shadows, lighting, and object occlusions.
\section{LiWi Framework}





\subsection{Shadow-Guided Learning}

\begin{figure*}[t]
    \centering
    \includegraphics[width=0.95\textwidth]{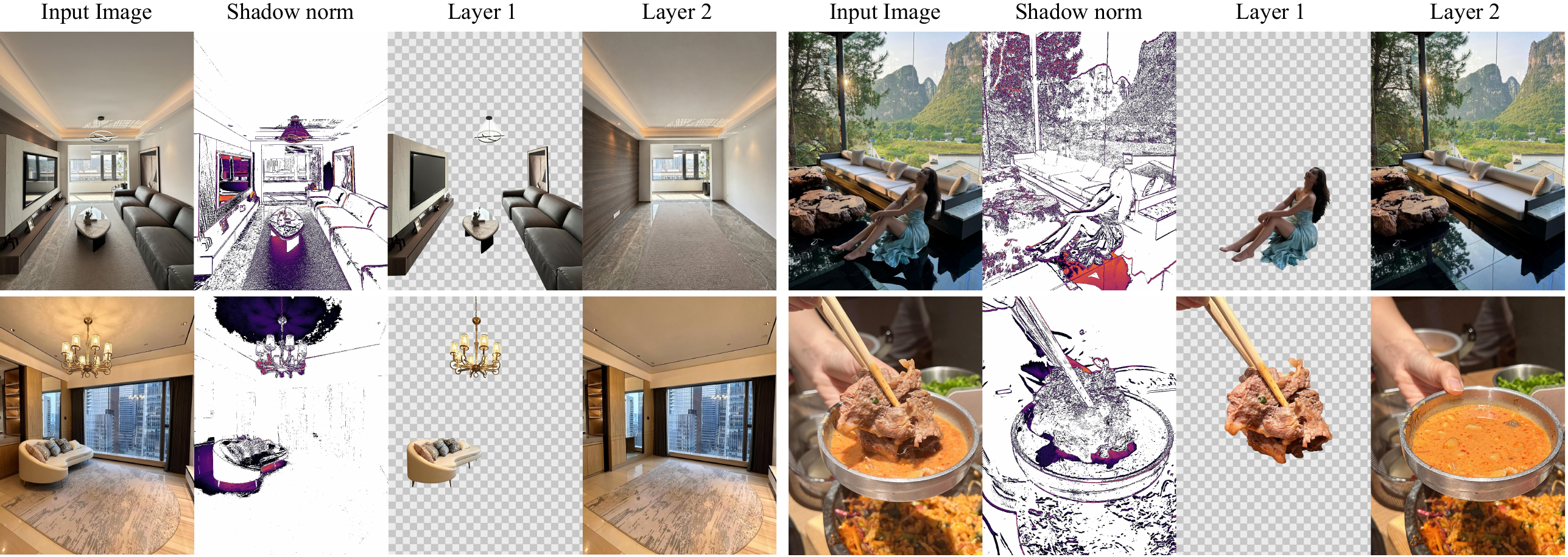}
    \caption{Effect of the shadow layer. The shadow layer records foreground-related lighting changes, such as shadows and occlusion, helping the model remove them when restoring a clean background.}
    \label{fig:method_shadow}
    \vspace{-10pt}
\end{figure*}

Real-world photographs contain complex photometric effects, such as cast shadows, illumination variations, and contact darkening.
As shown in \cref{fig:method_shadow}, we introduce a shadow layer to represent the footprint induced by foreground entities.
Specifically, let $I_{c}$ denote the recomposed image, which is obtained by stacking the background $B$ and all foreground layers $\{F_{k}\}_{k=1}^{K}$ in orders.
The shadow layer $S$ is defined as the residual between the source image $I_{src}$ and the recomposed image $I_{c}$, that is $S = I_{src} - I_{c}$.
Instead of forcing the illumination changes to be ambiguously absorbed by either the foreground or the background layer, we explicitly model the shadow layer.

\begin{figure}[htbp]
    \centering

    \begin{minipage}[b]{0.53\textwidth}
        \vspace{0pt}
        \centering
        \includegraphics[width=\linewidth]{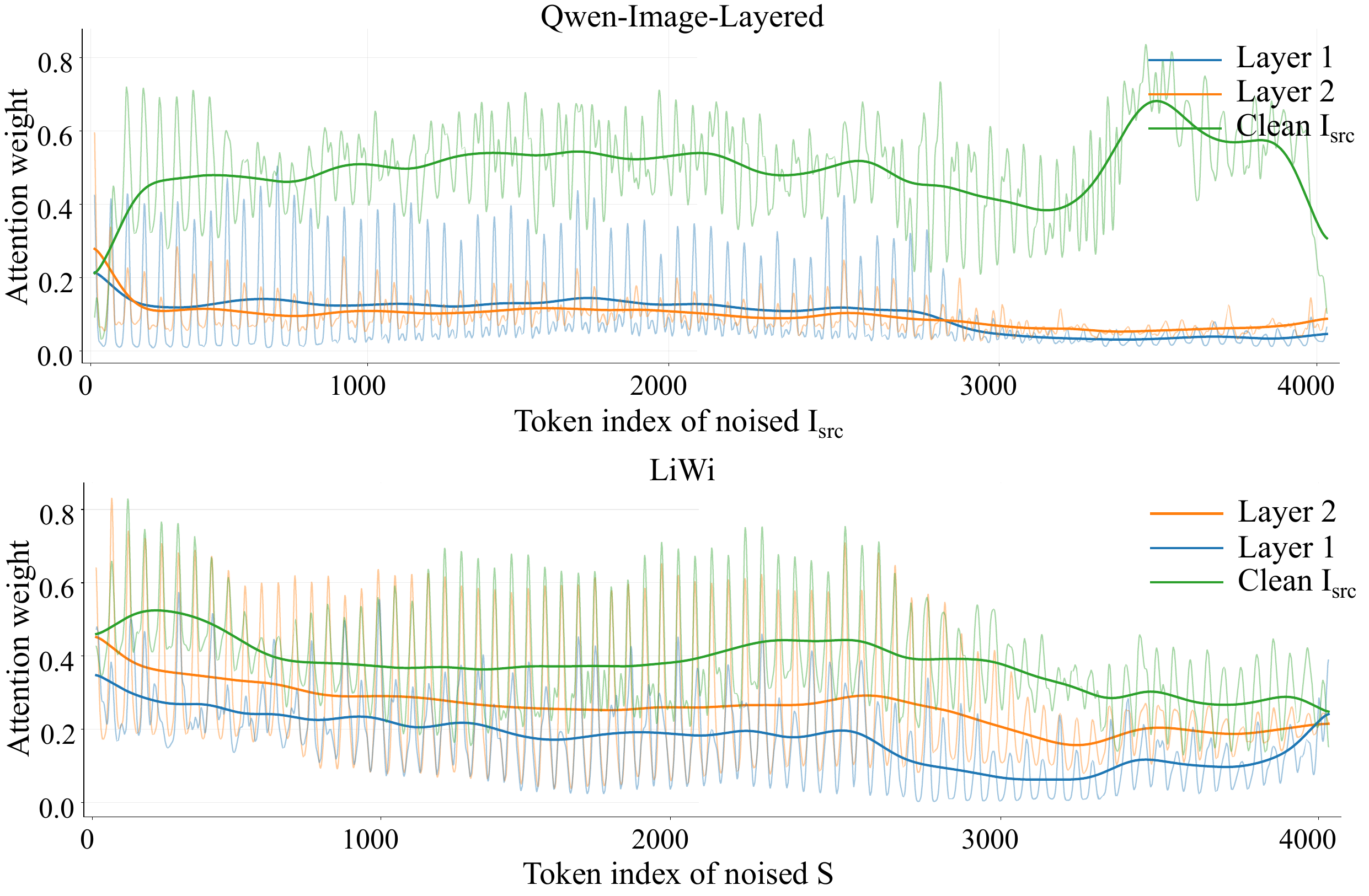}
        \caption{Attention-weight comparison between different reconstruction objectives.}
        \label{fig:method_attention}
    \end{minipage}
    \hfill
    \begin{minipage}[b]{0.35\textwidth}
        \vspace{0pt}
        \centering
        \includegraphics[width=\linewidth]{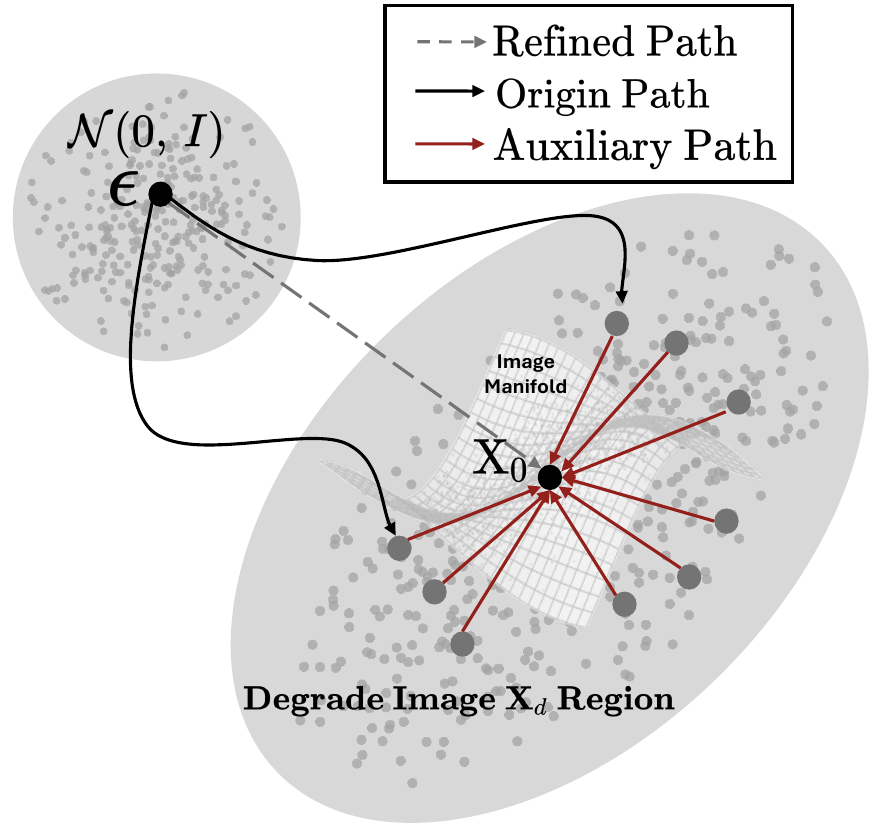}
        \caption{Illustration of the restoration process from degraded regions to the natural image manifold. }
        \label{fig:method_degradation}
    \end{minipage}

    \vspace{-10pt}
\end{figure}

During the training process of diffusion model, rather than regenerating the source image $I_{src}$ ~\cite{yin2025qwen}, we propose to model the generation process of shadow layer $S$, which avoids the arbitrary information propagation of artifacts.
To validate the effectiveness of this design, we analyze the attention weights of the noised $I_{src}$ or $S$ to other input tokens (\textit{i.e.}, clean $I_{src}$ and noised layers). 
When reconstructing $I_{src}$ (~\cref{fig:method_attention}, top), the model predominantly attends to the original input image. 
This may indicate the information leakage from clean reference image $I_{src}$ to noised $I_{src}$. 
When the objective is shifted to predicting $S$ (~\cref{fig:method_attention}, bottom), the attention distribution becomes more balanced among the clean $I_{src}$ and layered images.

By using the shadow layer to absorb the complex illumination variations, we prevent lighting artifacts from being erroneously attached to the layered images, and thus encourage layered images to concentrate on the generation of foreground entities and background scenes.
Therefore, the network successfully decouples the foreground entities and achieves improved accuracy in color consistency.

\subsection{Degraded Boundary Refinement}

In the layer generation task, given the ground-truth image $x_0 \in \{S\} \cup \mathcal{B} \cup \mathcal{F}$, the flow-matching~\cite{lipman2022flow} method constructs a linear path that transports a Gaussian sample $\epsilon$ to image $x_0$.
The latent representation at time step $t \in [0,1]$ is defined via linear interpolation:
\begin{equation}
    z_t=(1-t)\epsilon+t x_0 .
\end{equation}

However, natural images often contain complex structures of objects, leading to degraded boundary artifacts in the foreground generation. 
To refine the artifacts of generated foreground, we explicitly model the boundary refinement task in the diffusion model.
Specifically, we construct the degraded image $x_d$ from the ground-truth image $x_0 \in \mathcal{F}$ by erosion, dilation, or blurring of the boundary. 
an auxiliary flow path is introduced to transport the noised degradation image $x_d + \epsilon$ to the ground-truth image $x_0$, which yields the auxiliary path as:
\begin{equation}
z_t^{aux}=(1-t)(x_d+\epsilon)+t x_0 .
\end{equation}

As illustrated in \cref{fig:method_degradation}, we shift the start-point of Gaussian noise $\epsilon$ to the degraded observation $x_d + \epsilon$ around the ground-truth image $x_0$, expanding the exploration path for degraded boundary refinement.
This auxiliary path shares the same model weights with the original flow path in the foreground generation, and thus provides an additional supervision for boundary correction.
The final training objective combines the original flow matching loss and the auxiliary boundary-correction loss:

\begin{equation}
\begin{aligned}
\mathcal{L}
=&\ \mathbb{E}_{t,x_0 \in \{S\} \cup \mathcal{B} \cup \mathcal{F},\epsilon}
\left[
\left|v_\theta(z_t,t)-v_t\right|_2^2
\right] \
+\lambda
\mathbb{E}_{t,x_0 \in \mathcal{F},x_d,\epsilon}
\left[
\left|v_\theta(z_t^{aux},t)-v_t^{aux}\right|_2^2
\right],
\end{aligned}
\end{equation}

where $v_t = x_0 - \epsilon$, $v_t^{aux}=x_0-x_d-\epsilon$, and $\lambda$ controls the strength of auxiliary supervision.
In implementation, we update the attention mask and position embeddings due to the additional input of $x_d$.
For each degraded image, we use the same position embedding as its corresponding foreground image. 
In attention layer, each degraded image only attends to itself and the source image $I_{src}$. 
Therefore, the model learns both noise-to-image generation and degraded boundary correction, leading to more accurate boundaries.
During inference, we use the original flow path for layer generation, while the auxiliary path is only used as an additional training objective.




\section{Experiments}

\subsection{Experimental Setup}

\paragraph{Implementation Details.}
We train our model on the proposed LiWi-100k dataset, initializing the network with Qwen-Image-Layered \cite{yin2025qwen}.
The model is optimized using the Adam optimizer \cite{kingma2014adam} with a constant learning rate of $1 \times 10^{-5}$. 
Training is conducted on 16 NVIDIA B200 GPUs with a total batch size of 16 for 12K optimization steps. 
To efficiently process data with diverse structural layouts, we implement data bucketing strategy based on image aspect ratios and the number of layers. 
During training, the maximum image resolution is constrained within 1024×1024 pixels.

\paragraph{Datasets and Metrics.}
We evaluate our framework on two distinct benchmarks: our proposed LiWi-100k and the Crello \cite{yamaguchi2021canvasvae} test set. 
The LiWi-100k test set contains 1,000 in-the-wild images. 
In contrast, the Crello test set comprises 1,972 raster graphic design templates. 
Following LayerD \cite{suzuki2025layerd}, we report RGB L1 and Alpha soft IoU as the main evaluation metrics. 
RGB L1 measures the reconstruction accuracy of the predicted RGB layer appearance, where a lower value indicates better color and texture fidelity. 
Alpha soft IoU computes the IoU directly on the continuous alpha values, where a higher value indicates more accurate layer opacity and boundary estimation. 

\subsection{Quantitative Results}

\begin{table*}[h]
    \centering
    \vspace{-10px}
    \footnotesize 
    \setlength{\tabcolsep}{4pt} 
    
    \caption{Quantitative results on LiWi-100k test set. }
    \label{tab:reconstruction_quality}
     \scalebox{0.95}{
    \begin{tabular}{l|ccc|ccc}
        \toprule
        Metric & \multicolumn{3}{c|}{RGB L1 $\downarrow$} & \multicolumn{3}{c}{Alpha soft IoU $\uparrow$} \\
        \midrule
        \# Max Edits  & 0 & 1 & 2 & 0 & 1 & 2 \\
        \midrule
        
        Qwen-Image-Layered~\cite{yin2025qwen}
        & 0.2607 & 0.2581 & 0.2565
        & 0.6133 & 0.6187 & 0.6234 \\
        
        Qwen-Image-Layered-SFT
        & 0.0911 & 0.0908 & 0.0889
        & 0.9000 & 0.9009 & 0.9038 \\
        
        \textbf{LiWi}
        & \textbf{0.0822} & \textbf{0.0821} & \textbf{0.0810}
        & \textbf{0.9569} & \textbf{0.9574} & \textbf{0.9587} \\
        
        \bottomrule
    \end{tabular}}
    \vspace{-10px}
\end{table*}

\paragraph{Layer Decomposition.}
We report quantitative comparisons in \cref{tab:reconstruction_quality,tab:max_edits_quality}. On LiWi-100k, the original Qwen-Image-Layered model shows a clear domain gap when transferred from graphic designs to in-the-wild images. 
Qwen-Image-Layered-SFT which is fine-tuned on our data substantially improves both RGB reconstruction and alpha estimation. 
Compared with Qwen-Image-Layered-SFT, LiWi reduces RGB L1 by $9.41\%$ on average and improves alpha IoU by $6.22\%$. 
On the Crello~\cite{yamaguchi2021canvasvae} benchmark, LiWi also outperforms prior methods. 
Although Crello contains raster graphic designs rather than natural photographs, LiWi reduces RGB L1 by $12.45\%$ on average over Qwen-Image-Layered and improves alpha soft IoU by $1.35\%$. 
These results show that LiWi achieves strong gains on in-the-wild images while retaining robust performance on raster graphic designs.

\begin{table}[ht]
    \centering
    \vspace{-8px}
    \caption{Evaluation on Crello \cite{yamaguchi2021canvasvae} test set under different maximum edit numbers.}
    \label{tab:max_edits_quality}
    \footnotesize 
    \setlength{\tabcolsep}{2.5pt} 
    \scalebox{0.95}{
    \begin{tabular}{l|cccccc|cccccc}
        \toprule
        Metric & \multicolumn{6}{c|}{RGB L1 $\downarrow$} & \multicolumn{6}{c}{Alpha soft IoU $\uparrow$} \\
        \midrule
        \# Max Edits & 0 & 1 & 2 & 3 & 4 & 5 & 0 & 1 & 2 & 3 & 4 & 5 \\
        \midrule
        
        LayerD~\cite{suzuki2025layerd}
        & 0.0709 & 0.0541 & 0.0457 & 0.0419 & 0.0403 & 0.0396
        & 0.7520 & 0.8111 & 0.8435 & 0.8564 & 0.8622 & 0.8650 \\
        
        Qwen-Image-Layered~\cite{yin2025qwen}
        & 0.0594 & 0.0490 & 0.0393 & 0.0377 & 0.0364 & 0.0363
        & 0.8705 & 0.8863 & 0.9105 & 0.9121 & 0.9156 & 0.9160 \\
        
        \textbf{LiWi-Crello}
        & \textbf{0.0512} & \textbf{0.0423} & \textbf{0.0345} & \textbf{0.0331} & \textbf{0.0323} & \textbf{0.0321}
        & \textbf{0.8823} & \textbf{0.8956} & \textbf{0.9224} & \textbf{0.9250} & \textbf{0.9279} & \textbf{0.9308} \\
        \bottomrule
    \end{tabular}}
    \vspace{-8px}
\end{table}

\paragraph{Zero-Shot Foreground Segmentation.}
To further assess predicted alpha masks, we evaluate foreground segmentation on DIS-5K~\cite{DIS5K}, a high-resolution real-world benchmark with fine structures and diverse objects. 
As shown in ~\cref{tab:zero_shot_layered}, LiWi produces foreground masks competitive with specialized segmentation methods, despite having never seen these data. 
This suggests that our auxiliary boundary refinement helps capture subtle boundary cues.

\setlength{\extrarowheight}{2pt}
\begin{table*}[ht]
\centering
\vspace{-2px}
\caption{Comparison of various methods on the foreground segmentation.}
\label{tab:zero_shot_layered}
\footnotesize 
\setlength{\tabcolsep}{4pt} 
 \scalebox{0.95}{
\begin{tabular}{l|ccccc|ccccc}
\toprule
\multirow{2}{*}{Methods}
& \multicolumn{5}{c|}{TE (1-4)}
& \multicolumn{5}{c}{VD} \\
\cline{2-11}
& $F_\beta^x$ $\uparrow$ & $F_\beta^w$ $\uparrow$ & $\mathcal{M}$ $\downarrow$ & $S_m$ $\uparrow$ & $E_\phi^m$ $\uparrow$
& $F_\beta^x$ $\uparrow$ & $F_\beta^w$ $\uparrow$ & $\mathcal{M}$ $\downarrow$ & $S_m$ $\uparrow$ & $E_\phi^m$ $\uparrow$ \\
\midrule
BASNet~\cite{fan2020camouflaged}     & .744 & .664 & .092 & .786 & .814  & .737 & .656 & .094 & .781 & .809 \\
U$^2$Net~\cite{U-2-Net}              & .771 & .676 & .082 & .799 & .825  & .753 & .656 & .089 & .785 & .809 \\
HRNet~\cite{HRNet}                   & .743 & .658 & .087 & .781 & .840  & .726 & .641 & .095 & .767 & .824 \\
PGNet~\cite{PGNet}                   & .809 & .746 & .063 & .830 & .885  & .798 & .733 & .067 & .824 & .879 \\
IS-Net~\cite{DIS5K}                  & .799 & .726 & .070 & .819 & .858  & .791 & .717 & .074 & .813 & .856 \\
FP-DIS~\cite{FP-DIS}                 & .831 & .770 & .047 & .847 & .895  & .823 & .763 & .062 & .843 & .891 \\
BiRefNet~\cite{zheng2024bilateral}   & \textbf{.896} & \textbf{.858} & \textbf{.035} & \textbf{.901} & \textbf{.93}4  & \textbf{.891} & \textbf{.854} & \textbf{.038} & \textbf{.898} & \textbf{.931} \\
\midrule
\rowcolor{gray!15}
LiWi (Zero-shot)                          & .794 & .744 & .077 & .821 & .871  & .791 & .744 & .078 & .821 & .871 \\
\bottomrule
\end{tabular}
}
\vspace{-8px}
\end{table*}

\subsection{Qualitative Results}

\begin{figure*}[t]
    \centering
    \vspace{-10px}
    \includegraphics[width=0.92\textwidth]{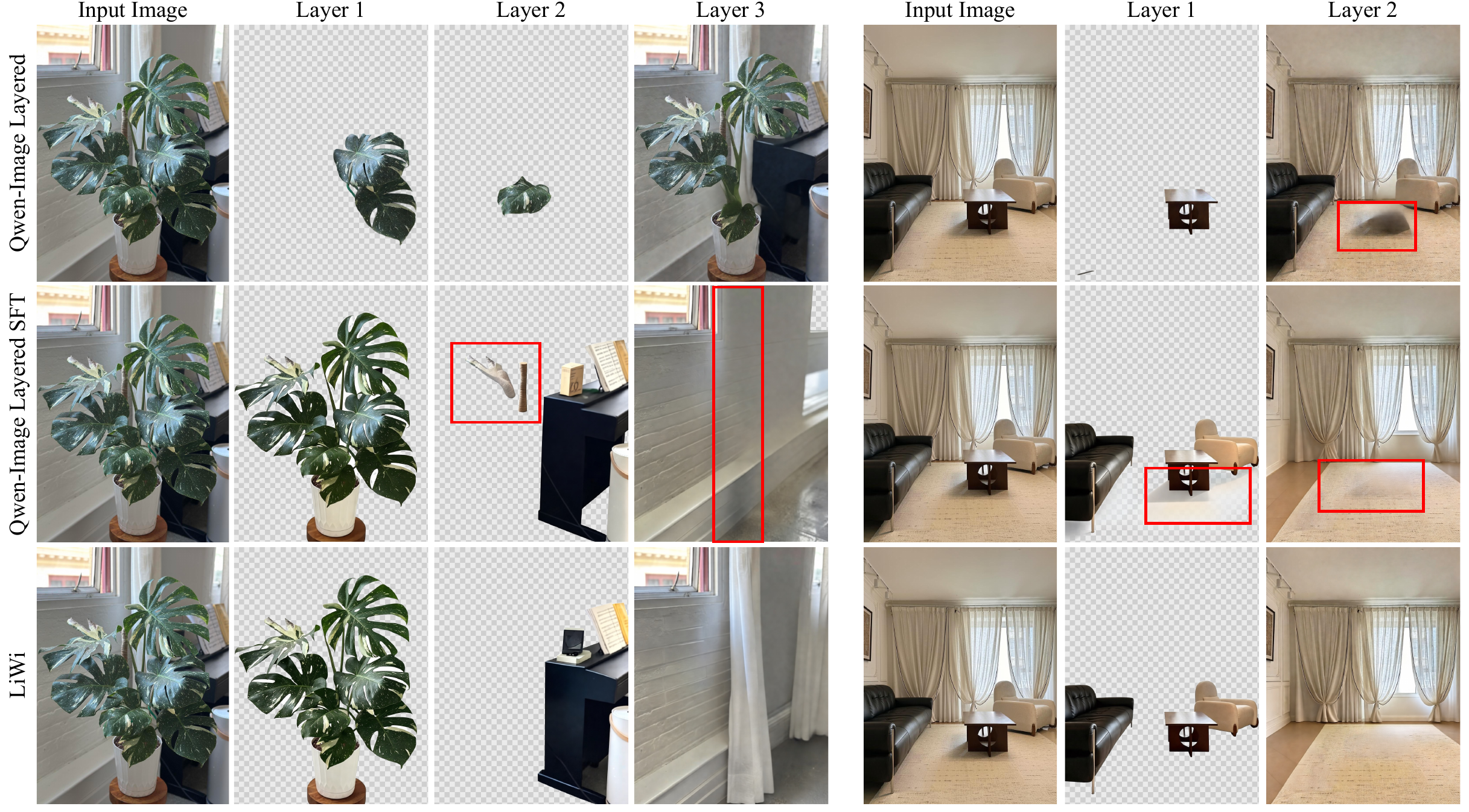}
    \caption{Qualitative comparison on in-the-wild layer decomposition. }
    \label{fig:exp_contrast}
\end{figure*}

\paragraph{Qualitative Layer Decomposition.}
Qualitative comparisons are presented in \cref{fig:exp_contrast}. LiWi consistently yields more faithful layered decompositions for in-the-wild images. In the plant example, Qwen-Image-Layered only extracts partial leaves, while the SFT variant introduces floating branch artifacts and erroneously removes background curtain structures. In contrast, LiWi preserves the foreground plant while maintaining background integrity. Furthermore, in indoor scenes where baselines leave residual contact shadows or dark regions indicating incomplete foreground-background disentanglement, LiWi effectively eliminates these artifacts to produce cleaner results.

\begin{figure*}[h]
    \centering
    \vspace{-10px}
    \includegraphics[width=0.92\textwidth]{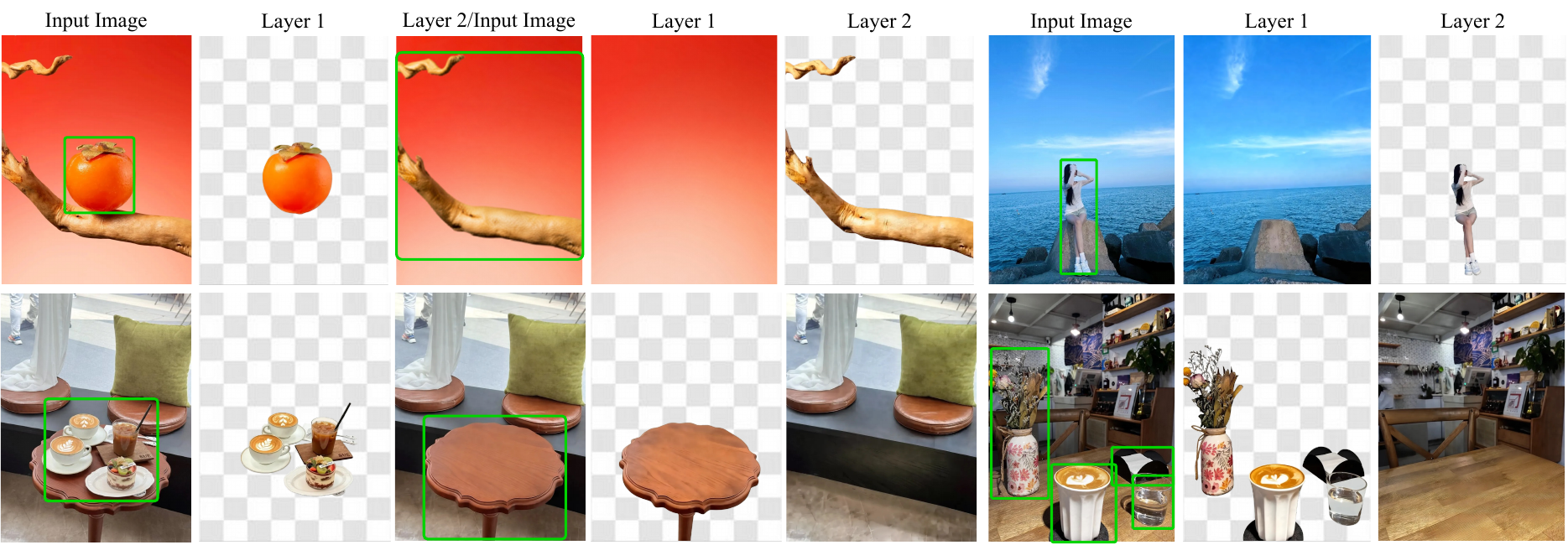}
    \caption{Layer decomposition guided by visual prompt.}
    \label{fig:vp}
    \vspace{-15px}
\end{figure*}

\paragraph{Visual Prompt for Layer Decomposition.}
To improve the generalizability and controllability of our method, we introduce visual-prompt-based layer decomposition. As shown in \cref{fig:vp}, given a user-specified bounding box that indicates the region to be separated, our model decomposes the corresponding content into an editable layer with an alpha mask. This process can be applied iteratively, enabling users to progressively decompose multiple regions in complex scenes.

\begin{table*}[h]
    \centering
    \vspace{-8px}
    \caption{Ablation of reconstruction targets and degradation-restoration objective on LiWi-100k.}
    \label{tab:shadow_horizontal}
    \footnotesize 
    \setlength{\tabcolsep}{4pt} 
     \scalebox{0.95}{
    \begin{tabular}{l|ccc|ccc}
        \toprule
        Metric & \multicolumn{3}{c|}{RGB L1$\downarrow$} & \multicolumn{3}{c}{Alpha soft IoU$\uparrow$} \\
        \midrule
        \# Max Edits  & 0 & 1 & 2 & 0 & 1 & 2 \\
        \midrule
        
        Qwen-Image-Layered-SFT
        & 0.0911 & 0.0908 & 0.0889
        & 0.9000 & 0.9009 & 0.9038 \\
        
        - Source Image Reconstruction
        & 0.0865 & 0.0866 & 0.0859
        & 0.9432 & 0.9454 & 0.9471 \\
        
        \quad + Latent-space Shadow
        & 0.0886 & 0.0886 & 0.0880
        & 0.9424 & 0.9433 & 0.9458 \\
        
        \quad + Pixel-space Shadow
        & 0.0824 & 0.0826 & 0.0817
        & 0.9499 & 0.9500 & 0.9522 \\

        \quad \quad + Degradation-restoration Objective
        & \textbf{0.0822} & \textbf{0.0821} & \textbf{0.0810}
        & \textbf{0.9569} & \textbf{0.9574} & \textbf{0.9587} \\
        
        \bottomrule
    \end{tabular}
    }
    \vspace{-8px}
\end{table*}

\subsection{Ablation Study}

We ablate the reconstruction targets in the layered diffusion objective in \cref{tab:shadow_horizontal}. 
Qwen-Image-Layered-SFT reconstructs the source image, creating a shortcut and allowing the model to bypass layer-wise reasoning. 
Removing this objective (- source image reconstruction ) leads to substantial improvements in both RGB L1 and Alpha IoU, indicating that direct image reconstruction weakens layer-level supervision.

We further investigate shadow supervision strategies. 
While latent-space shadow constructs shadows after encoding, pixel-space shadow forms them directly in image space. 
Among the two, pixel-space shadow performs better since shadow in latent space tend to be less expressive.
Compared with removing source reconstruction alone, pixel-space shadow further reduces RGB L1 by $4.75\%$ and improves alpha soft IoU by $0.58\%$ on average. 
This suggests that explicitly modeling shadow residuals helps the model better handle illumination-induced errors.

Finally, adding the degradation-restoration objective further improves alpha quality. It boosts alpha soft IoU by $0.73\%$  on average over pixel-space shadow, while yielding a smaller RGB L1 gain of $0.57\%$, consistent with its focus on refining boundary defects rather than global reconstruction.

\section{Conclusion and Limitations}

We presented LiWi, a framework for decomposing in-the-wild images.
To enable scalable supervision, we introduced ADD that automatically constructs in-the-wild layered images, resulting in LiWi-100k.
We further propose a novel framework for natural image layering.
The shadow layer captures illumination variations.
Meanwhile, the degradation-restoration objective provides auxiliary boundary-correction supervision.
Extensive experiments demonstrate that LiWi not only improves both RGB fidelity and alpha accuracy but also facilitates strong zero-shot foreground segmentation.

Despite these encouraging results, LiWi still has several limitations.
First, the quality of LiWi-100k depends on the capabilities of the agents, editing tools, segmentation experts, and verifiers.
Second, the selector-verifier design filters many unreliable samples, errors from object removal, mask estimation, or proposal verification may still be inherited by the final training data.
Future work may extend LiWi toward more physically grounded layer representations, stronger automatic data verification, and more complex multi-object real-world scenes.

\bibliographystyle{unsrtnat} 
\bibliography{refs}

\newpage
\appendix

\section{Complete Zero-Shot Foreground Segmentation Results on Real-World Data}
We compare our method with various foreground segmentation approaches on the four test sets and the validation set of DIS5K. The results show that our method achieves competitive performance under the zero-shot setting.
\begin{table*}[htbp]
\centering
\caption{Comparison of various methods on the foreground segmentation task. In the zero-shot setting, our method achieves performance close to that of dedicated foreground segmentation models.}
\label{tab:zero_shot_layered_full}

\resizebox{\textwidth}{!}{
\begin{tabular}{l|cccccc|cccccc}
\toprule
\multirow{2}{*}{Methods}
& \multicolumn{6}{c|}{TE1}
& \multicolumn{6}{c}{TE2} \\
\cline{2-13}
& $F_\beta^x$ $\uparrow$ & $F_\beta^w$ $\uparrow$ & $\mathcal{M}$ $\downarrow$ & $S_m$ $\uparrow$ & $E_\phi^m$ $\uparrow$ & $\text{HCE}_\gamma$ $\downarrow$
& $F_\beta^x$ $\uparrow$ & $F_\beta^w$ $\uparrow$ & $\mathcal{M}$ $\downarrow$ & $S_m$ $\uparrow$ & $E_\phi^m$ $\uparrow$ & $\text{HCE}_\gamma$ $\downarrow$ \\
\midrule
BASNet & .663 & .577 & .105 & .741 & .756 & 155 & .738 & .653 & .096 & .781 & .808 & 341 \\
U$^2$Net         & .701 & .601 & .085 & .762 & .783 & 165 & .768 & .676 & .083 & .798 & .825 & 367 \\
HRNet               & .668 & .579 & .088 & .742 & .797 & 262 & .747 & .664 & .087 & .784 & .840 & 555 \\
PGNet             & .754 & .680 & .067 & .800 & .848 & 162 & .807 & .743 & .065 & .833 & .880 & 375 \\
IS-Net             & .740 & .662 & .074 & .787 & .820 & 149 & .799 & .728 & .070 & .823 & .858 & 340 \\
FP-DIS             & .784 & .713 & .060 & .821 & .860 & 160 & .827 & .767 & .059 & .845 & .893 & 373 \\
UDUN                             & .784 & .720 & .059 & .817 & .864 & 140 & .829 & .768 & .058 & .843 & .886 & 325 \\
BiRefNet & .860 & .819 & .037 & .885 & .911 & 106 & .894 & .857 & .036 & .900 & .930 & 266 \\
\midrule
Ours                             & .812 & .755 & .064 & .836 & .880 & 204 & .825 & .777 & .065 & .846 & .891 & 505 \\
\bottomrule
\end{tabular}
}

\vspace{1mm}

\resizebox{\textwidth}{!}{
\begin{tabular}{l|cccccc|cccccc}
\toprule
\multirow{2}{*}{Methods}
& \multicolumn{6}{c|}{TE3}
& \multicolumn{6}{c}{TE4} \\
\cline{2-13}
& $F_\beta^x$ $\uparrow$ & $F_\beta^w$ $\uparrow$ & $\mathcal{M}$ $\downarrow$ & $S_m$ $\uparrow$ & $E_\phi^m$ $\uparrow$ & $\text{HCE}_\gamma$ $\downarrow$
& $F_\beta^x$ $\uparrow$ & $F_\beta^w$ $\uparrow$ & $\mathcal{M}$ $\downarrow$ & $S_m$ $\uparrow$ & $E_\phi^m$ $\uparrow$ & $\text{HCE}_\gamma$ $\downarrow$ \\
\midrule
BASNet & .790 & .714 & .080 & .816 & .848 & 681 & .785 & .713 & .087 & .806 & .844 & 2852 \\
U$^2$Net          & .813 & .721 & .073 & .823 & .856 & 738 & .800 & .707 & .085 & .814 & .837 & 2898 \\
HRNet              & .784 & .700 & .080 & .805 & .869 & 1049 & .772 & .687 & .092 & .792 & .854 & 3864 \\
PGNet               & .843 & .785 & .056 & .844 & .911 & 797 & .831 & .774 & .065 & .841 & .899 & 3361 \\
IS-Net             & .830 & .758 & .064 & .836 & .883 & 687 & .827 & .753 & .072 & .830 & .870 & 2888 \\
FP-DIS            & .868 & .811 & .049 & .871 & .922 & 780 & .846 & .788 & .061 & .852 & .906 & 3347 \\
UDUN                             & .865 & .809 & .050 & .865 & .917 & 658 & .846 & .792 & .059 & .849 & .901 & 2785 \\
BiRefNet & .925 & .893 & .028 & .919 & .955 & 569 & .904 & .864 & .039 & .900 & .939 & 2723 \\
\midrule
Ours                             & .809 & .760 & .068 & .833 & .887 & 1026 & .727 & .685 & .108 & .767 & .822 & 3700 \\
\bottomrule
\end{tabular}
}

\vspace{1mm}

\resizebox{\textwidth}{!}{
\begin{tabular}{l|cccccc|cccccc}
\toprule
\multirow{2}{*}{Methods}
& \multicolumn{6}{c|}{TE (1-4)}
& \multicolumn{6}{c}{VD} \\
\cline{2-13}
& $F_\beta^x$ $\uparrow$ & $F_\beta^w$ $\uparrow$ & $\mathcal{M}$ $\downarrow$ & $S_m$ $\uparrow$ & $E_\phi^m$ $\uparrow$ & $\text{HCE}_\gamma$ $\downarrow$
& $F_\beta^x$ $\uparrow$ & $F_\beta^w$ $\uparrow$ & $\mathcal{M}$ $\downarrow$ & $S_m$ $\uparrow$ & $E_\phi^m$ $\uparrow$ & $\text{HCE}_\gamma$ $\downarrow$ \\
\midrule
BASNet & .744 & .664 & .092 & .786 & .814 & 1007 & .737 & .656 & .094 & .781 & .809 & 1132 \\
U$^2$Net          & .771 & .676 & .082 & .799 & .825 & 1042 & .753 & .656 & .089 & .785 & .809 & 1139 \\
HRNet               & .743 & .658 & .087 & .781 & .840 & 1432 & .726 & .641 & .095 & .767 & .824 & 1560 \\
PGNet               & .809 & .746 & .063 & .830 & .885 & 1173 & .798 & .733 & .067 & .824 & .879 & 1326 \\
IS-Net              & .799 & .726 & .070 & .819 & .858 & 1016 & .791 & .717 & .074 & .813 & .856 & 1116 \\
FP-DIS             & .831 & .770 & .047 & .847 & .895 & 1165 & .823 & .763 & .062 & .843 & .891 & 1309 \\
UDUN                             & .831 & .772 & .057 & .844 & .892 & 977 & .823 & .763 & .059 & .838 & .892 & 1097 \\
BiRefNet & .896 & .858 & .035 & .901 & .934 & 916 & .891 & .854 & .038 & .898 & .931 & 989 \\
\midrule
Ours                             & .794 & .744 & .077 & .821 & .871 & 1359 & .791 & .744 & .078 & .821 & .871 & 1417 \\
\bottomrule
\end{tabular}
}
\end{table*}

\section{Visualization Results of the Auxiliary Path}
\begin{figure}[htbp]
    \centering
    \includegraphics[width=\textwidth]{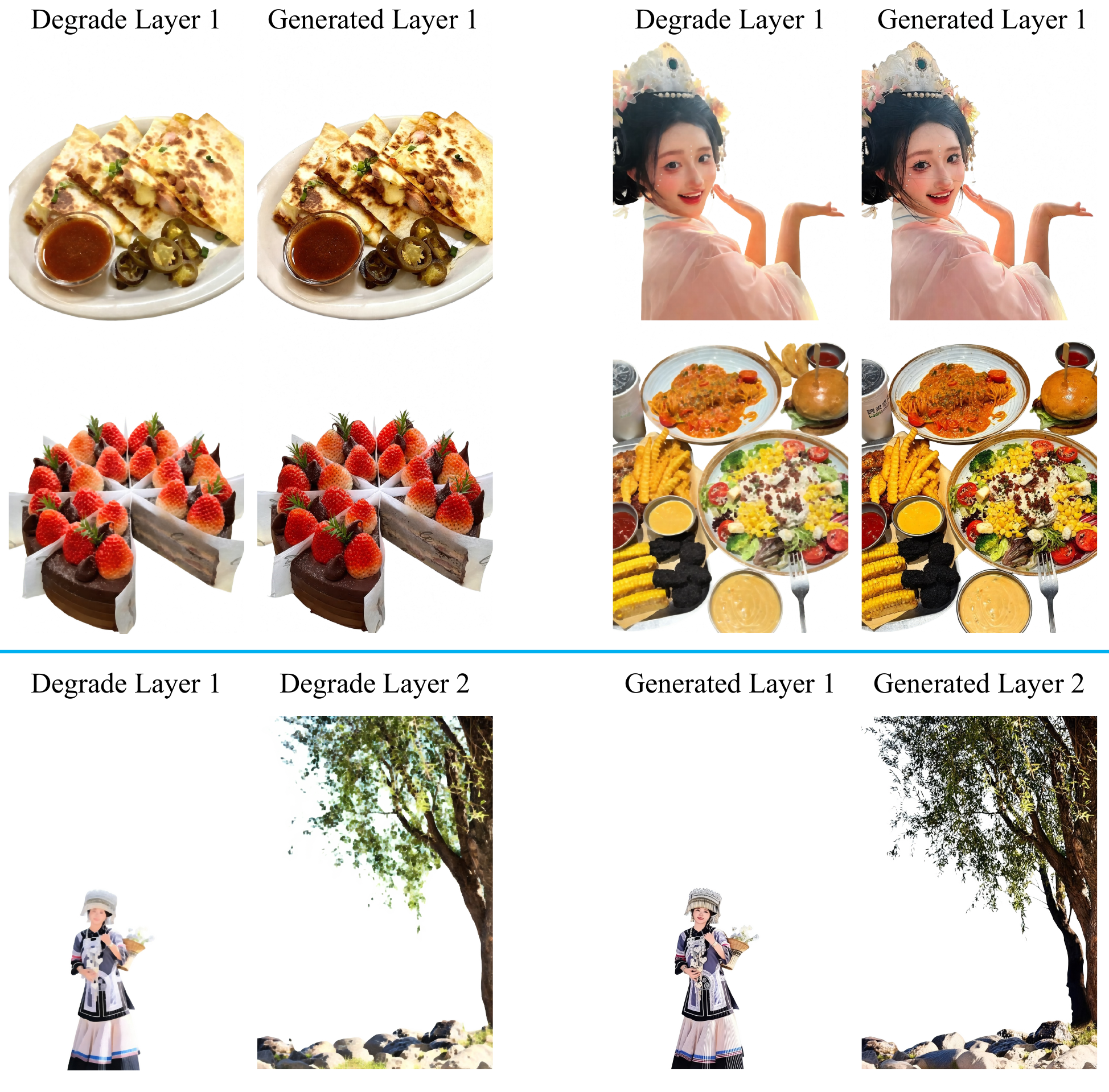}
    \caption{The degraded layer is obtained by expanding the original image region and then applying erosion. The decomposed layer is generated from this degraded layer through the auxiliary path. As shown in the results, our auxiliary path can effectively recover plausible layer content from the eroded input.}
    \label{fig:app_degrade}
\end{figure}

To further illustrate the effect of the auxiliary path, we present some visualization results in ~\cref{fig:app_degrade}. The degraded layer is obtained by first expanding the original image region and then applying erosion, while the decomposed layer is generated from this degraded input through the auxiliary path. As shown in the figure, our auxiliary path can effectively recover plausible layer content from eroded inputs, demonstrating its ability to refine degraded boundaries and restore coherent layer structures.

\section{Additional visualization Results}
\subsection{Additional visualization Results generated on LiWi framework}
\begin{figure}[htbp]
    \centering
    \includegraphics[width=\textwidth]{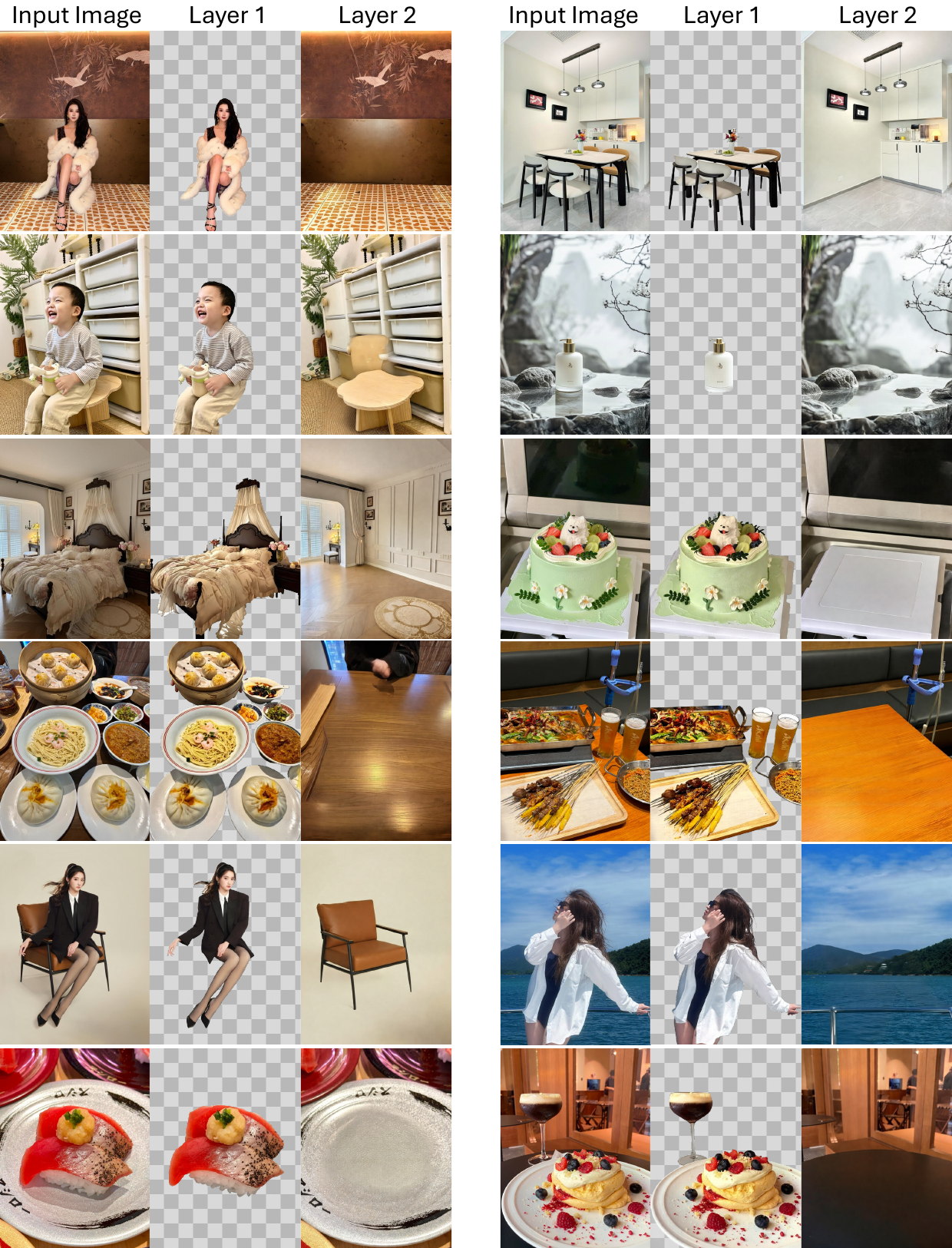}
    \caption{Results of LiWi framework on the test set of LiWi-100k. For various natural scenes with multiple categories and diverse illumination conditions, our method can perform high-quality layering while maintaining the consistency of light and shadow.}
    \label{fig:visulization_method}
\end{figure}

To further demonstrate the generative capabilities of our proposed framework, we provide additional image samples generated on the test set of LiWi-100k. It can be seen in Fig.~\ref{fig:visulization_method} that our method can accomplish layered tasks in diverse scenarios, including simultaneous layering of multiple objects, portrait layering, layering under specific lighting conditions, etc. While completing layering with high quality, our method maintains the consistency of lighting and shadow.

\subsection{Additional visualization Results on LiWi-100k dataset}
\begin{figure}[htbp]
    \centering
    \includegraphics[width=\textwidth]{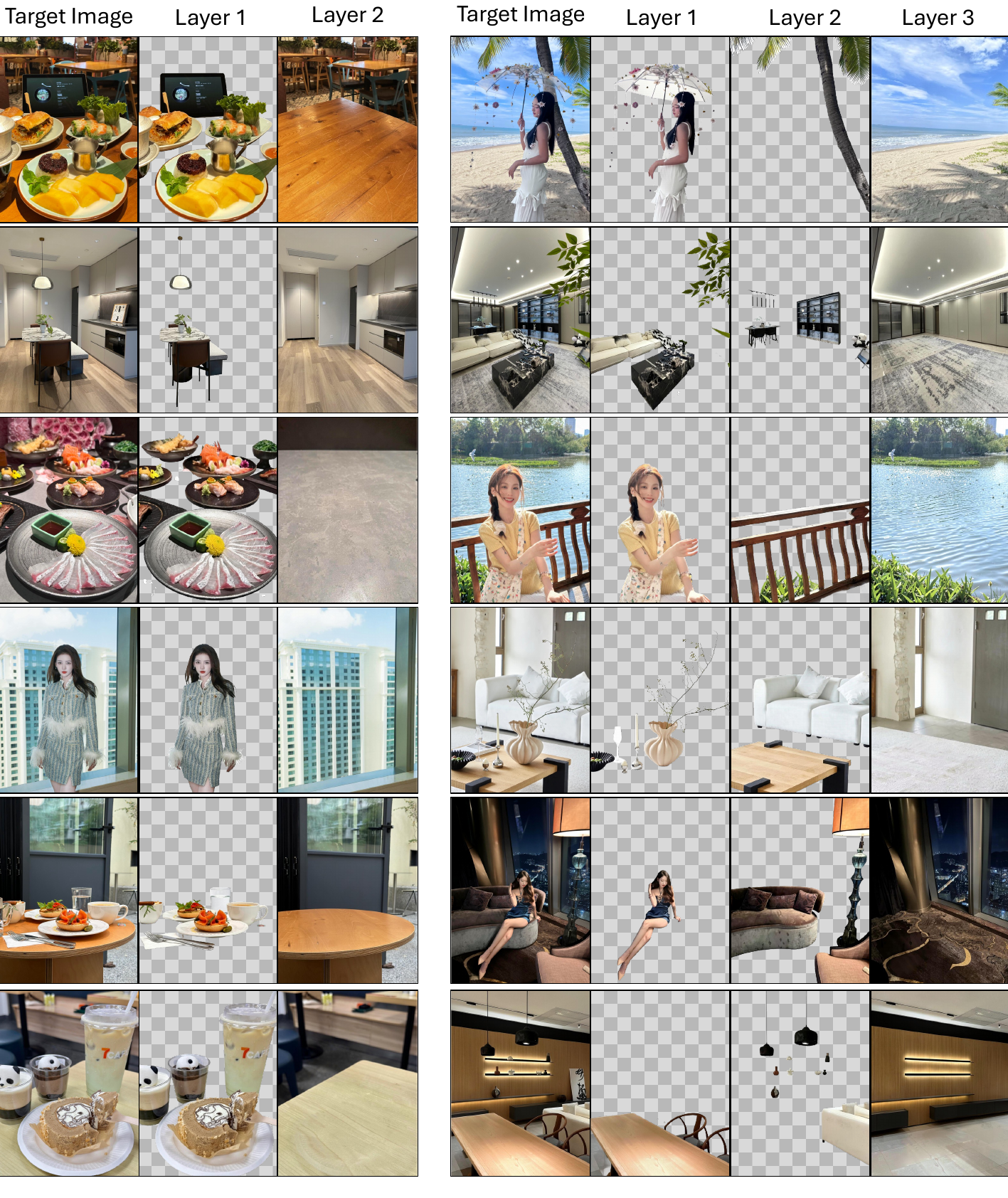}
    \caption{Visualization of the Liwi dataset with 2 and 3 layers. As shown, in diverse scenes, our construction method can generate high-quality layered images.}
    \label{fig:visulization_dataset_23}
\end{figure}

\begin{figure}[htbp]
    \centering
    \includegraphics[width=\textwidth]{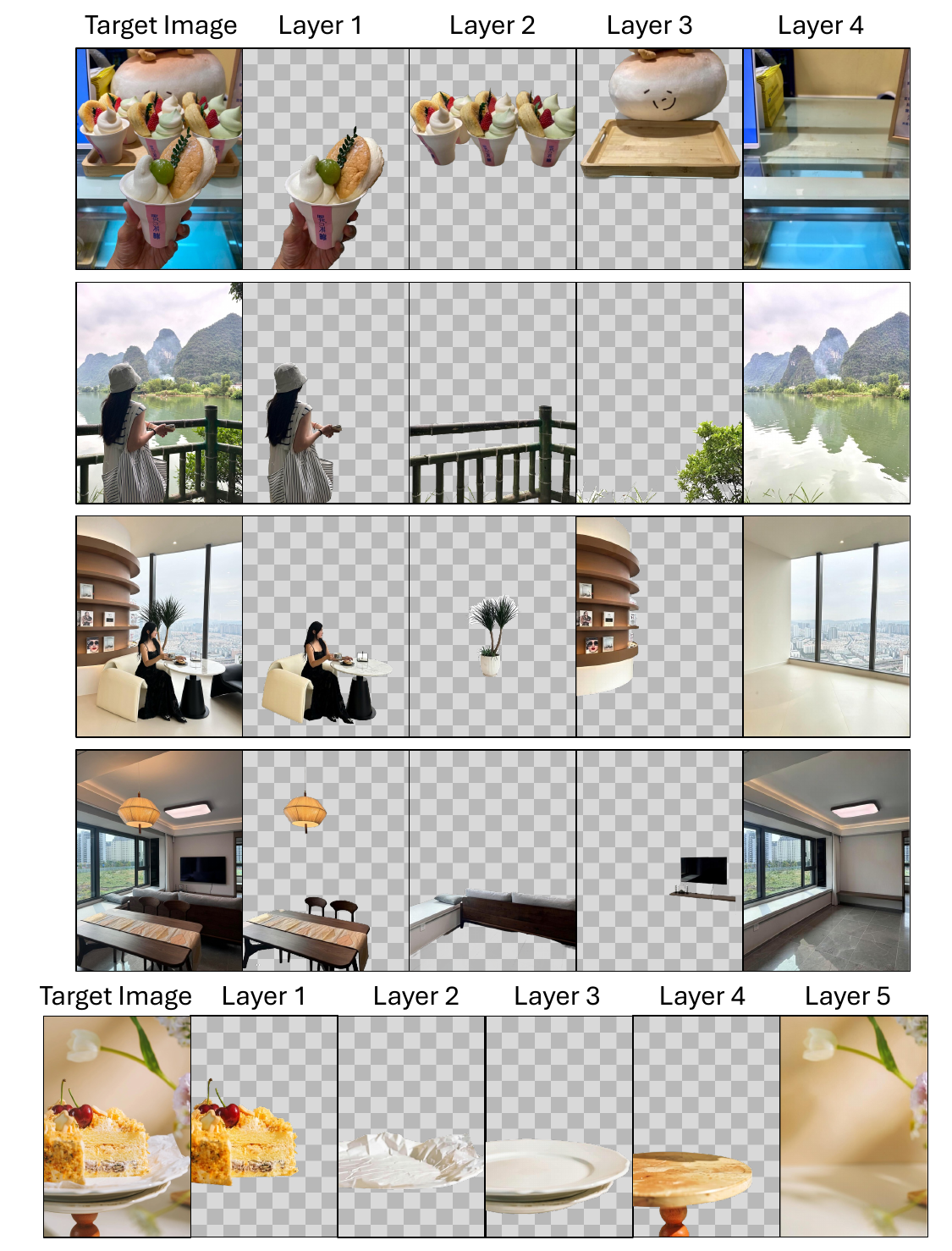}
    \caption{Visualization of the LiWi-100k dataset across multiple layers and aspect ratios. As the number of layers increases and the aspect ratio changes, our dataset construction method can still produce high-quality hierarchical results with semantic meaning.}
    \label{fig:visulization_dataset_45}
\end{figure}

To further demonstrate the diversity of the dataset, we present additional image samples from LiWi-100k. As shown in the Fig.~\ref{fig:visulization_dataset_23} and Fig.~\ref{fig:visulization_dataset_45}, across both diverse scene categories and semantically rich multi-level decomposition tasks, our dataset construction method can consistently generate high-quality hierarchical results with high accuracy. While performing semantic decomposition, it also preserves the consistency and richness of lighting and shadow effects.






\end{document}